\documentclass{article}

\usepackage{microtype}
\usepackage{graphicx}
\usepackage{subfigure}
\usepackage{booktabs} 

\usepackage{listings}
\usepackage[nolist]{acronym}

\begin{acronym} 
\acro{RL}{Reinforcement Learning}
\acro{HER}{Hindsight Experience Replay}
\acro{DR}{Domain Randomization}
\acro{PER}{Prioritized Experience Replay}
\acro{TD}{Temporal Difference}
\acro{MC}{Monte Carlo}
\end{acronym}

\usepackage{hyperref}

\usepackage[accepted]{icml2021}

\icmltitlerunning{Karolos: An Open-Source Simulation and Reinforcement Learning Suite}

\begin{document}

\twocolumn[
\icmltitle{Karolos: An Open-Source Reinforcement Learning Framework for Robot-Task Environments}



\icmlsetsymbol{equal}{*}

\begin{icmlauthorlist}
\icmlauthor{Christian Bitter}{equal,tmdt}
\icmlauthor{Timo Thun}{equal,tmdt}
\icmlauthor{Tobias Meisen}{tmdt}
\end{icmlauthorlist}

\icmlaffiliation{tmdt}{Institute for Technologies and Management of the Digital Transformation, University of Wuppertal, Wuppertal, Germany}

\icmlcorrespondingauthor{Christian Bitter}{bitter@uni-wuppertal.de}

\icmlkeywords{Reinforcement Learning, Industrial Robotics}

\vskip 0.3in
]



\printAffiliationsAndNotice{\icmlEqualContribution} 

\begin{abstract}
In \ac{RL} research, simulations enable benchmarks between algorithms, as well as prototyping and hyper-parameter tuning of agents.
In order to promote \ac{RL} both in research and real-world applications, frameworks are required which are on the one hand efficient in terms of running experiments as fast as possible. On the other hand, they must be flexible enough to allow the integration of newly developed optimization techniques, e.g. new \ac{RL} algorithms, which are continuously put forward by an active research community.
In this paper, we introduce \emph{Karolos}, a \ac{RL} framework developed for robotic applications, with a particular focus on transfer scenarios with varying robot-task combinations reflected in a modular environment architecture. In addition, we provide implementations of state-of-the-art \ac{RL} algorithms along with common learning-facilitating enhancements, as well as an architecture to parallelize environments across multiple processes to significantly speed up experiments.
The code is open source and published on GitHub with the aim of promoting research of \ac{RL} applications in robotics.
\end{abstract}

\section{Introduction}

Reinforcement Learning (RL) aims at learning to solve tasks by an agent exploring an environment in a trial-and-error fashion. An agent observes the state of an environment and decides for an action according to its policy. Performing the action results in a state change of the environment. By receiving rewards that reflect the status of a certain task and gaining sufficient experience, an agent can derive which actions are useful with respect to solving the task. Recent successes of reinforcement learning include learning strategies for games such as Go \cite{Silver.2017} and StarCraft \cite{Vinyals.2019}, to industrial settings such as production scheduling \cite{kemmerling2021towards} and robotic manipulation tasks \cite{Levine.2016, meyes2018continuous}.

In real-world robotics scenarios, the highly iterative, data-driven nature of reinforcement learning has led to the use of simulations to pre-train agents and subsequently perform a sim2real transfer to reduce real-world training efforts \cite{Tobin.2017}. On top of easing access to environments, thus facilitating benchmarking, simulations run faster and can be parallelized, which significantly speeds up experiments. Also, actions in simulations are consequence-free, meaning that performing sub-optimal actions does not bear the risk of hardware damage and associated costs.

As \ac{RL} is an active field of research with first real-world applications, two major domains can be identified. On the one hand, research on new algorithms and techniques to make learning more efficient is mostly focused on improving agents. Hereby, to effectively compare different approaches, it is essential that the benchmark environments, such as OpenAI Gym \cite{Brockman.05.06.2016}, exist and can be utilized. However, research by its very nature also necessitates a higher degree of flexibility and modifiability of software to model particular aspects not relevant to most other use cases. On the other hand, applications of \ac{RL} are focused on framing a certain use-case as an environment and adapting existing and tested algorithms by matching interfaces and setting hyper-parameters to the problem at hand. While both domains, research and application of \ac{RL}, focus on different aspects, they naturally depend on each other and benefit from similar optimization techniques.



In this work, we introduce our software framework \emph{Karolos} aimed at research of \ac{RL} in the robotics domain with a distinct focus on cross-robot transfer learning. Transfer learning in general is concerned with the reuse of knowledge acquired in one domain towards facilitating learning in another domain. Cross-robot transfer learning as a subfield of research is aimed at transferring knowledge about solving a common task between different robotic systems, which may differ in both morphology and/or abilities. Our contribution for this field of research is a \ac{RL} framework which is
\begin{itemize}
    \item \textbf{modular}: In order to accommodate a wide spectrum of application scenarios, a generalized framework must allow for a high degree of flexibility regarding the combination as well as parametrization of modular components, such as \ac{RL} algorithms and environments. Additionally, as \ac{RL} is an active field of research, it is important to keep frameworks flexible enough to incorporate future optimization techniques developed by the community.
    In order to facilitate research into cross-robot transfer learning, we provide a modular infrastructure which allows the composition of environments by mix-and-matching robot models with tasks. However, it is also possible to quickly integrate other (non-robotic) \ac{RL} environments, such as the gym benchmark \cite{Brockman.05.06.2016}.
    \item \textbf{fast}: \ac{RL} experiments are usually very data-intensive due to their trial-and-error nature. As applications become more complex, the need for speeding up experiments to keep costs feasible will only increase. Besides improving the efficiency of learning algorithms, it is also possible to use multiple environments in parallel to generate more interaction experience for an agent to learn from.
    In \emph{Karolos}, we provide common optimization techniques, such as \ac{PER} \cite{Schaul.18.11.2015}, \ac{HER} \cite{Andrychowicz.05.07.2017} and \ac{DR} \cite{Tobin.2017}. In addition, our architecture allows massive parallelization of environments.
    \item \textbf{open-source}: In order to facilitate research, it is important to utilize software packages accessible to everyone. The framework is built using open-source software components in order to ensure access, regular updates and long-term support \cite{paulson2004empirical}. To allow operation without the need to acquire additional software licenses, \emph{Karolos} is based on the open-source physics engine Pybullet \cite{ErwinCoumans.20162019} and open-source tensor computation library Pytorch \cite{Paszke.2019}.
\end{itemize}

The framework is available at \url{https://github.com/tmdt-buw/karolos}.

\section{Software description}

In the following we describe the software architecture of our framework \textit{Karolos} and motivate the various design choices.

\subsection{Software Architecture}

\begin{figure} 
\includegraphics[width=\linewidth]{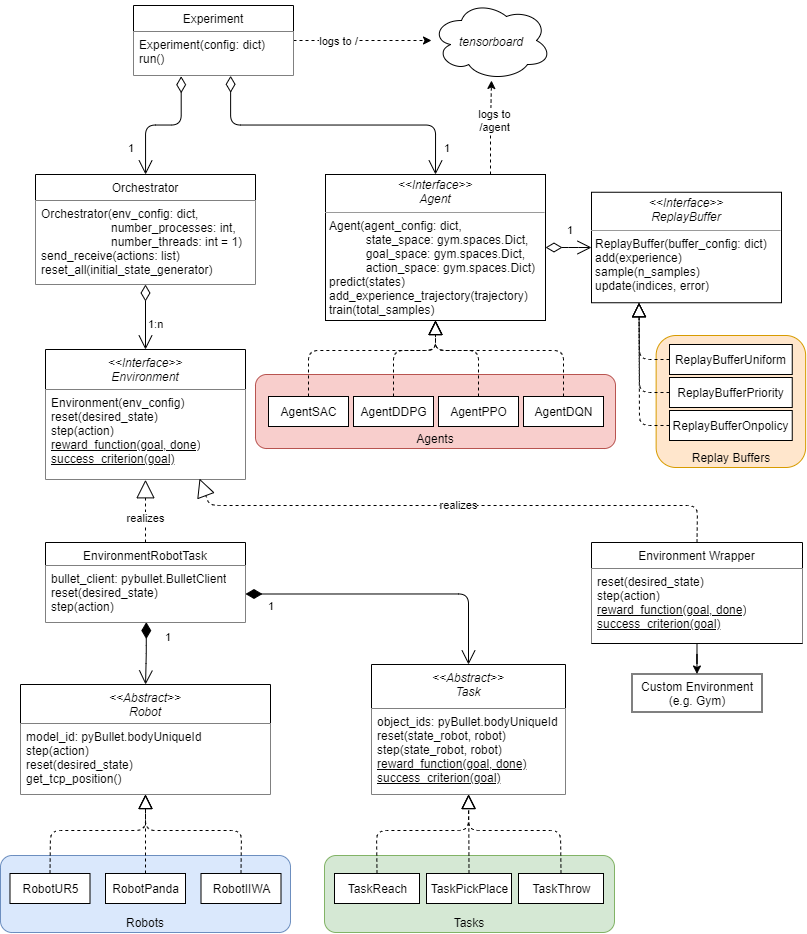}
\caption{The software architecture of \textit{Karolos} consists of Environments, which are combinations of Robots and Tasks, and Agents used to learn a respective task. The training is controlled and monitored by the Manager.}
\label{fig:architecture}
\end{figure}

The architecture of the \emph{Karolos} framework is depicted in Figure \ref{fig:architecture}. When launching an experiment, the \textbf{Experiment} serves as the entry point. The class Experiment coordinates the interactions between an \ac{RL} agent and one or multiple environments. Its main functions are to keep track of the episode count and monitor agent performance by testing at regular intervals, meaning that the agent is executed without exploration for a specified amount of epochs to calculate performance metrics. An experiment can be monitored in real-time, as all parameters are logged using tensorboard \cite{MartnAbadi.2015}.

The collection of experience is one major bottleneck when prototyping \ac{RL} agents. For this purpose, the class \textbf{Orchestrator} is utilized to parallelize multiple copies of an environment. Parallelization is hereby possible by multiprocessing, multithreading or a combination of both. The individual environments are addressed via unique IDs, which are used by an instance of the Experiment class to match asynchronously collected corresponding states and actions into valid experience trajectories.

An instance of the interface \textbf{Environment} represents an individual environment and provides an interface which allows resetting and executing actions. In addition, each environment must explicitly define a reward function and success criterion. The information required to determine reward and success is not included in the state, but separated in a goal variable. These design choices are required for data augmentation methods such as \ac{HER} \cite{Andrychowicz.05.07.2017}, which retroactively adjust experience based on hypothetical goals.

\begin{figure} 
\includegraphics[width=\linewidth]{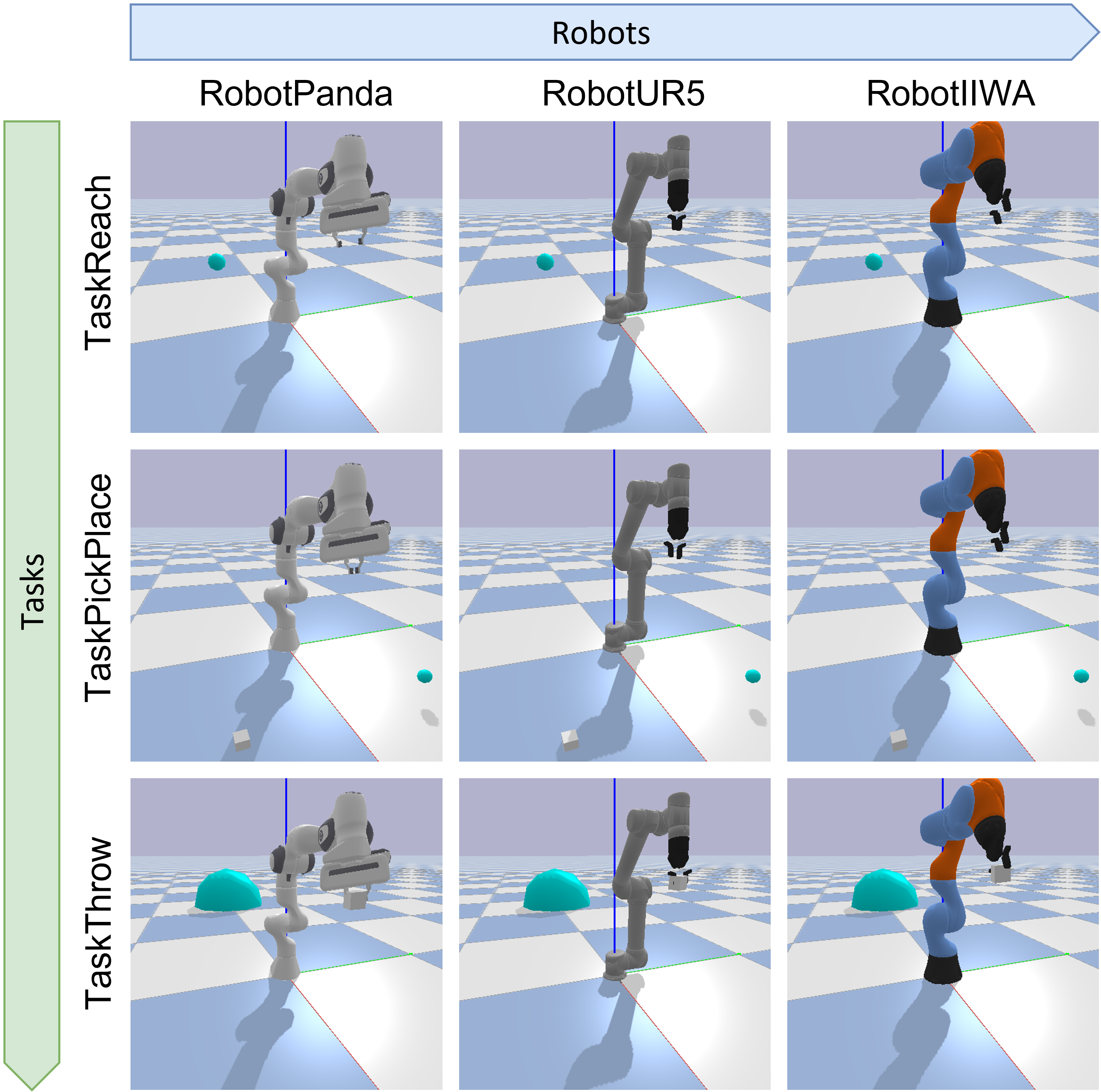}
\caption{The separation of an environment into robots and tasks allows for arbitrary combination into a multitude of robotic scenarios.}
\label{fig:robot_task}
\end{figure}

As \emph{Karolos} is developed with a focus on research into cross-robot transfer learning \cite{Devin.2017, Gupta.2017, Fickinger.07.10.2021}, a composable environment \textbf{EnvironmentRobotTask} is provided. The environment is further structured into a \textbf{Robot} and a \textbf{Task}, which can be combined arbitrarily as demonstrated in Figure \ref{fig:robot_task}. The Robot represents the physical unit which can be manipulated by an agent. It's step function therefore takes an action and translates it into movement. The Task contains all elements which constitute the environment surrounding the robot, such as manipulable objects and obstacles. It also provides the reward function and success criterion, which formalize the task at hand. Thus We utilize Pybullet as the physics engine to simulate the interactions in the environment \cite{ErwinCoumans.20162019}. 

To date, three robotic arms are available: a Franka Emika Panda \footnote{www.franka.de}, a Universal Robots UR5 \footnote{www.universal-robots.com} and a KUKA LBR iiwa \footnote{www.kuka.com}, whereby the latter two are equipped with a Robotiq 2F-85 Gripper \footnote{www.robotiq.com}. The robots were chosen, as they are in general attractive both for research and industrial applications due to them being safer and cheaper than traditional industrial robots \cite{galin2020cobots}. As the robots are self-contained modules, adding new robots and using them interchangeably is straightforward.

Complementary, three tasks are implemented:
\begin{itemize}
    \item Reach: A robot's end effector must be moved to a specified target position. This task arguably poses the simplest robotic task of reaching a desired configuration, without interaction with any objects or obstacles.
    \item Pick\_Place: An object must be picked up and moved to a desired target location. This task poses a more realistic challenge, as pick-and-place scenarios are common applications for industrial robots \cite{dillmann2020types}.
    \item Throw: An object is spawned in the robot's gripper. The robot must be moved and gripper opened so that the released object reaches a target location outside the robot's reach. This scenario introduces a delayed reward, as the agent must learn that it's success only depend on actions performed prior to the release of the object.
\end{itemize}

Analogous to robots, tasks are stand-alone modules which facilitates the implementation of new tasks and allows for arbitrary robot-task combinations.

Finally, an instance of the \textbf{Agent} interface contains the actual \ac{RL} algorithm. 
It hosts the models required for training and making predictions, and uses a \textbf{Replay Buffer} to store and sample collected experience. 
To date, four deep \ac{RL} architectures are implemented in \emph{Karolos}: soft actor-critic (SAC) \cite{Haarnoja.13.12.2018}, deep deterministic policy gradient (DDPG) \cite{Lillicrap.10.09.2015}, deep Q-learning (DQN) \cite{double-dqn} and Proximal Policy Optimization (PPO) \cite{ppo}. 

\subsection{Additional Optimization Techniques}

In this section, we highlight three popular concepts which are implemented as extensions in \emph{Karolos}: \ac{PER} (\ref{per})
and \ac{HER} (\ref{her}) which facilitate off-policy algorithmic convergence and \ac{DR} (\ref{dr}) to bridge the simulation-to-reality gap for transferring algorithms pre-trained in simulation to real-world setups. As before, modularity was emphasized and led to numerous important design choices, which will be explained in the following.

\subsubsection{Prioritized Experience Replay}
\label{per}

During training, not all experience is equally valuable with regard to improving an agent. It is much more efficient to show the agent samples of sparse events rather than samples which occur frequently in order to maximize the insights to be gained. 
One option to achieve this is by using a \ac{PER} buffer \cite{Schaul.18.11.2015}. 
Instead of sampling data from the pool of collected experience uniformly at every learning step, samples are weighted according to the magnitude of the \ac{TD} error.
The weights of the individual samples can hereby be determined by a probability distribution, which formalizes the deviation of the state-action valuation by the agent to the actual valuation under consideration of the observed reward. 
By doing so, the agent effectively optimizes on samples more frequently where it's value predictions are still off.
In \emph{Karolos} \ac{PER} is implemented in the class \textbf{ReplayBufferPriority}. By modularizing the replay buffer functionality it is possible to switch to other sampling strategies, e.g. an first-in-first-out sampling without replacement strategy (\textbf{ReplayBufferOnpolicy}) used in on-policy algorithms such as PPO. 

\subsubsection{Hindsight Experience Replay}
\label{her}
Finding a dense reward function which guides the agent into the direction of a solution can be a hard endeavour. Instead, many complex tasks can only be described by a sparse reward function. This, however, impedes learning as the chances of an agent receiving a positive reward signal requires it to solve the task during exploration, the probability of which diminishes exponentially with the amount of steps required for solving a task. \ac{HER} is a strategy to mitigate the implications of sparse reward functions on policy optimization \cite{Andrychowicz.05.07.2017}. The intuition behind \ac{HER} is to learn as much from a failed attempt as from a successful one. This is achieved by modifying an experience trajectory after completion. In essence, the states in the experiences are modified in such a way that the goal is no longer the original one, but instead corresponds to the state the agent ended up in. This data augmentation provides the agent with experience where the goal was achieved, even if it was missed in the actual trajectory.
In order to achieve \ac{HER} in \emph{Karolos}, experience is buffered by the instance of Experiment until an episode is terminated. The complete trajectory is then handed over to the agent, which generates new samples according to \ac{HER} and finally stores the original and the new \ac{HER} experience samples in its memory buffer.

\subsubsection{Domain Randomization}
 \label{dr}
The final goal of robotics research is to deploy agents to the real world, with simulations allowing for a fast and economic pretraining. However, the inherent complexity resulting from unknown or complicated causalities in the real world makes exact simulations infeasible. \ac{DR} is an approach to overcome this gap between simulation and reality \cite{Tobin.2017}. 
By intentionally introducing parameter modifications into training, an agent is exposed to many environment variations. 
Doing so increases the robustness of an agent to such variations, which increases the chances of a successful transfer to the real world. In \emph{Karolos}, parameters such as gravity or joint frictions are sampled from parametrizable normal distributions at each reset of an environment. The randomization is thereby delegated to individual modules, i.e. the randomization of robot parameters is handled by the robot instance.

\section{Illustrative Examples}
\label{}

Launching an experiment with \emph{Karolos} can be achieved with a few lines of code, as depected in Listing \ref{code:minimal}. Each experiment is specified by a dictionary, which is stored in the results directory for documentation. After launching the experiment, the training progress is logged into the results directory and can be monitored in real time using tensorboard.

\begin{lstlisting}[language=Python, caption=Minimal example of launching an experiment in Karolos, label={code:minimal}]

from karolos.experiment import Experiment

Experiment({
    "total_timesteps": 5_000_000,
    "test_interval": 500_000,
    "agent_config": {
        "name": "sac",
    },
    "env_config": {
        "robot_config": {
            "name": "panda",
        },
        "task_config": {
            "name": "reach",
        },
    }
}).run(results_dir="./results")

\end{lstlisting}

\subsection{Agent Training}

\begin{figure}[!htbp]
\centering
\begin{minipage}[t]{0.475\textwidth}
\includegraphics[width=\textwidth]{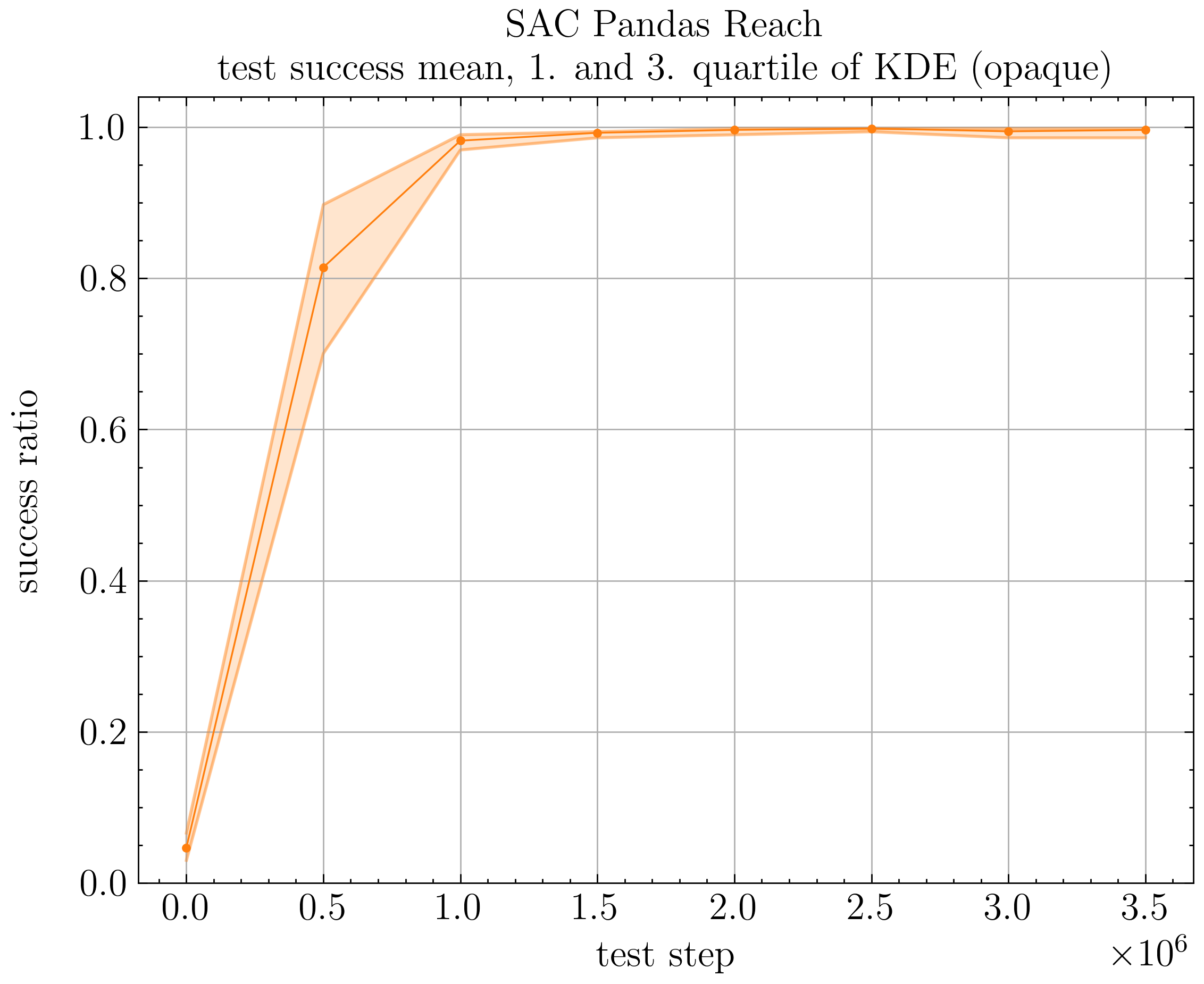}
\caption{Test success ratio of solving the task TaskReach with the robot RobotPanda using the agent AgentSAC.}
\label{fig:sac}
\end{minipage}
\hfill
\begin{minipage}[t]{0.475\textwidth}
\includegraphics[width=\textwidth]{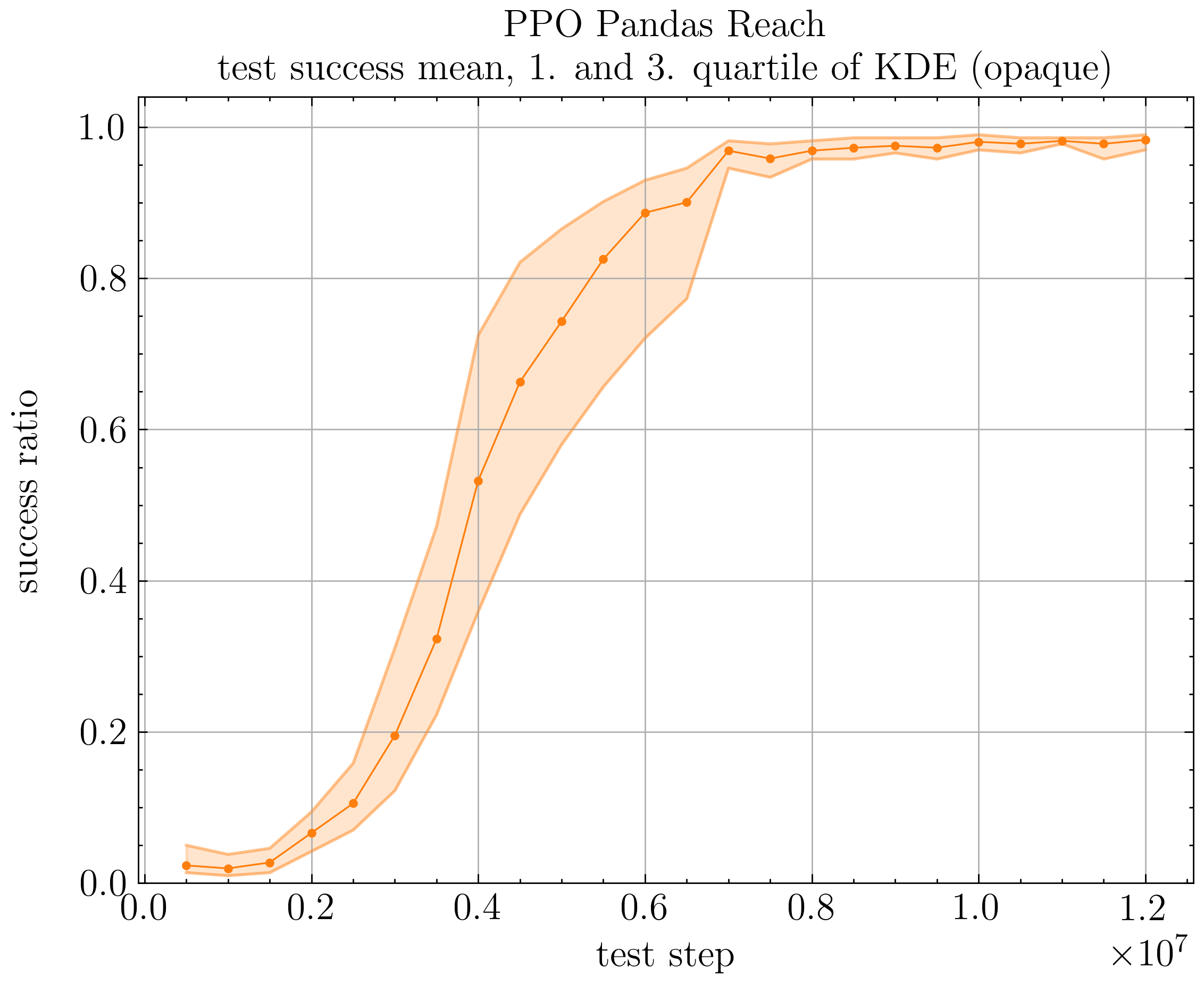}
\caption{Test success ratio of solving the task TaskReach with the robot RobotPanda using the agent AgentPPO.}
\label{fig:ppo}
\end{minipage}
\end{figure}

To demonstrate general functionality of \emph{Karolos}, we train two agents, SAC and PPO on solving the task TaskReach with the robot RobotPanda. The reach task serves as good benchmark for general functionality, as it is the simplest robotics task. The resulting success ratios are displayed in Figures \ref{fig:sac} and \ref{fig:ppo}. The success ratio measures the percentage of cases in which the success\_criterion of the task was met during a certain number, in our case 100, test episodes with random starting conditions using the policy of the agent. Each algorithm is executed seven times with different random seeds, and a kernel density estimation (KDE) technique using Simpson's rule is applied to approximate the underlying probability density function of each algorithms' performance. The results clearly show, that the reach task can be learned with the implementations in \emph{Karolos}.

\subsubsection{Parallelization}

\begin{figure}[!htbp]
\centering
\begin{minipage}[t]{0.475\textwidth}
\includegraphics[width=\textwidth]{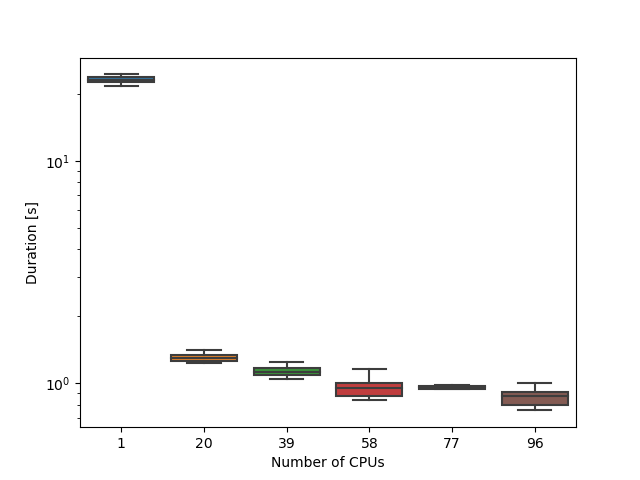}
\caption{Benchmarking experiment of wall clock speed increase by spawning environments on multiple CPUs. Each CPU holds one environment.}
\label{fig:cpu}
\end{minipage}
\hfill
\begin{minipage}[t]{0.475\textwidth}
\includegraphics[width=\textwidth]{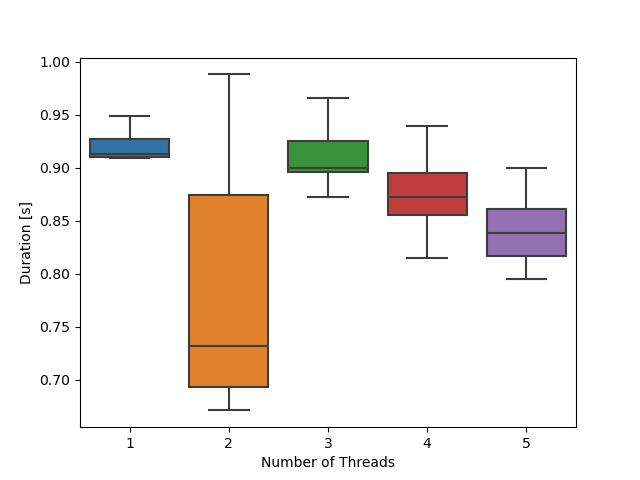}
\caption{Further speed improvement by spawning multiple environments per CPU. A total of 96 CPUs was used.}
\label{fig:thread}
\end{minipage}
\end{figure}

As described above, one central entity of \emph{Karolos} is the Orchestrator, which is used to spawn environments in parallel in multiple processes and threads. To showcase the speedup achieved by parallelization, we perform two benchmarking experiments. In the first experiment, we initialize Orchestrators to use an increasing amount of CPUs. For each configuration, environments are reset 1.000 times and the wall clock time is recorded. Each configuration is run 10 times and the results are displayed in Figure \ref{fig:cpu}. As expected, the durations significantly decrease with increasing parallelization through multiprocessing. While not as extensive, similar observations can be made when increasing the number of environments even further with multithreading, as depicted in Figure \ref{fig:thread}. Here, we chose the maximum number of CPUs available (96), as this setting yielded the best performance in the previous experiment. The benchmarking code is available in the examples folder and may be used to determine the optimal orchestrator parameters for a given system.

\subsection{Industrial Applications}

In addition to its application in academic research and education of reinforcement learning for robotics, \emph{Karolos} was already successfully utilized in two industrial research settings, which we will describe in the following.

\subsubsection{Use-Case 1: Robotic Assembly in Airplane Manufacturing}

\begin{figure}[!htbp]
\centering
\begin{minipage}[t]{0.475\textwidth}
\includegraphics[width=\textwidth]{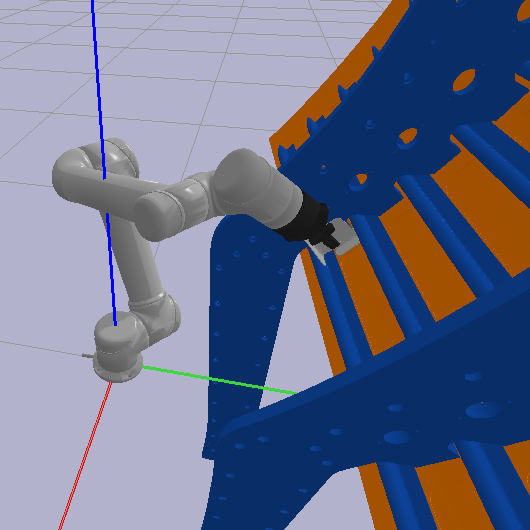}
\end{minipage}
\hfill
\begin{minipage}[t]{0.475\textwidth}
\includegraphics[width=\textwidth]{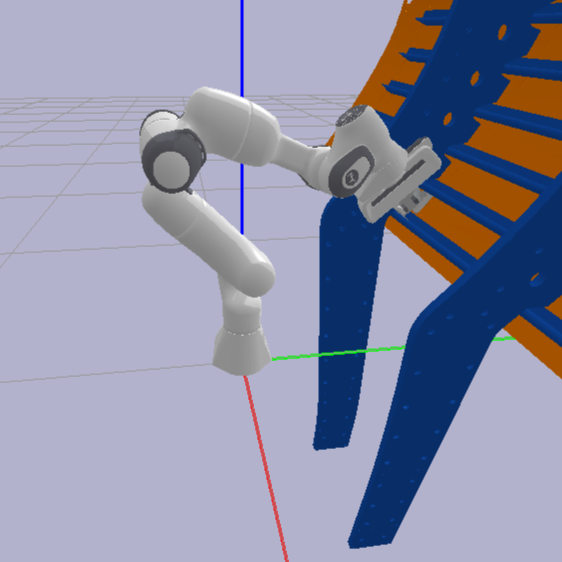}
\end{minipage}
\caption{The simulation environment developed for research into the assembly of aircraft shells with different collaborative robots.}
\label{fig:agreed}
\end{figure}

In the aerospace industry, many assembly tasks are still performed manually due to two reasons: On the one hand, product quantities are relatively low compared to other industries, such as automotive. Thus, amortization periods are much longer. On the other hand, structures are usually very big and difficult to move, which means that a robotic solution needs to be mobile, which brings additional challenges, such as the compensation of offsets resulting from movement. To address these challenges, a research project was conducted to investigate the possibilities for employing \ac{RL} as means to reduce manual programming efforts and thus make an automation solution more viable. In addition, the cross-robot transfer necessary to enable an interchangeability of different cobots was investigated \ref{fig:agreed}. \emph{Karolos} was chosen as backbone for the \ac{RL} architecture. Effectively, we were able to reuse the entire training pipeline as-is. The only major modification lay in the creation of a suitable environment on the basis of process specifications and proprietary virtual models, which were integrated into a simulation based on Pybullet. The simulation environment is depicted in Figure \ref{fig:agreed}. The experiments, and the success of the project in direct dependence, significantly benefited from the ability of parallelization and the resulting reduction of time-to-insights.

\subsubsection{Use-Case 2: Order-based Job Shop Scheduling}

In an array of industries such as lumber \cite{faaland1984log}, glass \cite{durak2017dynamic} or textile \cite{ozdamar2000cutting}, the task lies in cutting a big piece of material into smaller chunks. Specifically, the yield is to be increased by cutting around defects in a way which produces chunks to certain quality specifications. To investigate the possibility to employ \ac{RL} as means to learn this optimization problem under consideration of boundary conditions, such as different order sizes and quality requirements, an industrial research project was conducted. Again, \emph{Karolos} was chosen as \ac{RL} framework. In this instance, the availability of a pre-existing process simulation required the implementation of a suitable environment wrapper, after which the rest of the learning architecture was reusable to conduct necessary experiments. While not surprising due to the domain-agnostic nature of \ac{RL}, the ability to significantly reduce the ramp-up to the necessary experiments hints towards a usability of \emph{Karolos} for use-cases in domains different from our initial focus on robotics.

\section{Related Work}

In recent years numerous software packages emerged in the \ac{RL} domain, which focus on different aspects of research - from building environments suitable for benchmarking purposes to implementing the algorithms themselves in order to facilitate application of \ac{RL}.

The most similar projects to \emph{Karolos} in terms of the robotic environment are panda-gym \cite{gallouedec2021pandagym} and robosuite \cite{robosuite2020}, which both share similar design decisions regarding the separation of robots and tasks. The major differences to \emph{Karolos} are, that panda-gym only provides an implementation of the Panda robot and robosuite is based on the Mujoco physics engine \cite{E.Todorov.2012}. Other projects, such as rl-reach \cite{aumjaud2021rl_reach}, Gymperium \cite{BenjaminEllenberger.}, Gym \cite{Brockman.05.06.2016} or Metaworld \cite{Yu.24.10.2019} do not provide the flexibility in composing scenarios by selecting both the robot and the task.

One important sidenote regarding projects based on Mujoco \cite{E.Todorov.2012} is the circumstance that the python package mujoco-py \footnote{https://github.com/openai/mujoco-py} was deprecated for Windows. While Mujoco was aquired by the company Deepmind and the new Python package mujoco supports Windows, to date many projects have not been updated accordingly. Being built upon Pybullet, \emph{Karolos} can be run on Windows, Linux and Mac machines.

Due to the generalized and straight-forward formalization of \ac{RL}, the implementation of our training pipeline shares many similarities, and is partly inspired by, other projects \cite{raffin2019stable, PabloSamuelCastro.2018, Gauci.2018, Liang.26.12.2017}, with rlpyt \cite{Stooke.04.09.2019} being the most similar to our implementation in \emph{Karolos} regarding the architectural structure. Two aspects which set \emph{Karolos} apart are on the one hand its utilization of multithreading for further parallelization of environments. While the most significant speedup is achieved through multiprocessing, the minor improvement by multithreading scales, as the tasks \ac{RL} is applied to become more and more complex. On the other hand, the modular architecture allows an easy way to implement or customize agents. Some projects either prioritize a lean code base, such as stable-baselines \cite{raffin2019stable}, which is aimed at benchmarking different \ac{RL} algorithms and exhibits a pronounced inheritance structure. Other projects, such as rlpyt \cite{Stooke.04.09.2019} provide a degree of modularity, however in some cases not sufficiently. One example of this are is \ac{PER}, which is implemented for the deep-q agent (DQN) and thus not available for the other agents without reimplementation.

\section{Conclusions and Future Work}

This paper presents our reinforcement-learning framework \emph{Karolos}, with which we provide environments of robotic and common \ac{RL} benchmarking tasks as well as \ac{RL} algorithms. 
We deliberately chose open-source components for \emph{Karolos}, as we see a necessity to provide open-source training frameworks to the research community. 
In addition, we emphasized modularity in the software architecture, which facilitates the integration of new components and concepts into the framework.

While to date the framework is functional and can be used to train agents successfully, the project is under active development, which is driven by our applications in both academic and industrial research projects.
Our focus in the near future will be set on the expansion of the available simulations to create a benchmark for robot-to-robot transfer learning \cite{Devin.2017, Gupta.2017} as a complement to research into task-to-task transfer learning \cite{Yu.24.10.2019}. We also plan on using the foundations laid in this work to to continuously expand \emph{Karolos} by new robots, tasks and agents.

\bibliography{main.bib}
\bibliographystyle{icml2021}

\end{document}